\documentclass[journal]{IEEEtran}
\IEEEoverridecommandlockouts
\usepackage{cite}
\usepackage{balance}
\usepackage{amsmath,amssymb,amsfonts}
\usepackage{algorithmic}
\usepackage{graphicx}
\usepackage{textcomp}
\usepackage{xcolor}
\usepackage{booktabs}
\usepackage{subfigure}
\usepackage{multirow}
\usepackage{xcolor}
\usepackage{soul}
\sethlcolor{pink}
\usepackage{tablefootnote}
\usepackage[ruled,vlined]{algorithm2e}
\usepackage[utf8]{inputenc}

\usepackage{amsthm}

\def \x {\mathbf{x}}
\def \W {\mathbf{W}}

\def\BibTeX{{\rm B\kern-.05em{\sc i\kern-.025em b}\kern-.08em
    T\kern-.1667em\lower.7ex\hbox{E}\kern-.125emX}}
    
\begin{document}
\markboth{Journal of Biomedical and Health Informatics}%
{Shell \MakeLowercase{\textit{et al.}}: Bare Demo of IEEEtran.cls for IEEE Journals}

\title{Personalized On-Device E-health Analytics with Decentralized Block Coordinate Descent
}

\author{Guanhua~Ye, Hongzhi~Yin, Tong~Chen, Miao~Xu, Quoc~Viet~Hung~Nguyen, and Jiangning~Song
        
\thanks{
G. Ye, H. Yin, T. Chen and X. Miao are with the School of Information Technology \& Electric Engineering, The University of Queensland, Australia. E-mail: g.ye@uq.edu.au, h.yin1@uq.edu.au, tong.chen@uq.edu.au, miao.xu@uq.edu.au. 

Q.V.H. Nguyen is with the School of Information and Communication Technology, Griffith University. E-mail: henry.nguyen@griffith.edu.au.

J. Song is with the School of Biomedical Sciences and Monash Biomedicine Discovery Institute, Monash University. E-mail: jiangning.song@monash.edu. 
}
\thanks{H. Yin is the corresponding author.}
}%

\maketitle

\begin{abstract}
Actuated by the growing attention to personal healthcare and the pandemic, the popularity of E-health is proliferating. Nowadays, enhancement on medical diagnosis via machine learning models has been highly effective in many aspects of e-health analytics. Nevertheless, in the classic cloud-based/centralized e-health paradigms, all the data will be centrally stored on the server to facilitate model training, which inevitably incurs privacy concerns and high time delay. Distributed solutions like Decentralized Stochastic Gradient Descent (D-SGD) are proposed to provide safe and timely diagnostic results based on personal devices. However, methods like D-SGD are subject to the gradient vanishing issue and usually proceed slowly at the early training stage, thereby impeding the effectiveness and efficiency of training. In addition, existing methods are prone to learning models that are biased towards users with dense data, compromising the fairness when providing E-health analytics for minority groups. 

In this paper, we propose a Decentralized Block Coordinate Descent (D-BCD) learning framework that can better optimize deep neural network-based models distributed on decentralized devices for E-health analytics. As a gradient-free optimization method, Block Coordinate Descent (BCD) mitigates the gradient vanishing issue and converges faster at the early stage compared with the conventional gradient-based optimization. To overcome the potential data scarcity issues for users' local data, we propose similarity-based model aggregation that allows each on-device model to leverage knowledge from similar neighbor models, so as to achieve both personalization and high accuracy for the learned models. Benchmarking experiments on three real-world datasets illustrate the effectiveness and practicality of our proposed D-BCD, where additional simulation study showcases the strong applicability of D-BCD in real-life E-health scenarios.
\end{abstract}
 

\IEEEpeerreviewmaketitle

\section{Introduction} \label{sec:intro}
\IEEEPARstart{E}{-health} enables patients to receive medical advice and services without traveling to medical facilities~\cite{telemedicine}. Propelled by the advances in telecommunication technologies and the ongoing pandemic, many clinicians started offering E-health services for the first time in 2020 \cite{telemedicinetrend}. Except for necessary medical examinations and surgeries, the prosperity of E-health is revolutionizing people's stereotype about visiting a doctor. As an ongoing trend, enhancement upon E-health using machine learning (ML) models, especially deep neural networks (DNNs) has been shown highly effective in a wide range of E-health predictive analytics, such as automated diagnosis \cite{sun2020disease}, respiratory disease monitoring \cite{ye2021fenet}, and medication recommendation \cite{chen2020try}.

In the classic paradigm of enhancing remote healthcare with ML, the data collected from the patient (e.g., personal information, medical records, health sensor data, etc.) needs to be transferred to the cloud, i.e., a central server that provides all the computing resources for model training and inference \cite{baker2017internet}. In this cloud-based setting, all the sensitive user data is uploaded and stored within the central server. Consequently, this incurs high vulnerability to adversarial attacks that threaten both the security of user data (e.g., attribute inference attacks \cite{mosallanezhad2019deep}) and the usability of the trained E-health models (e.g., false data injection \cite{koh2017understanding}). Recently, the increasing number of severe privacy breaches in E-health systems is challenging the public's confidence in the security of such fully centralized services, such as the 2020 E-health malware attack in Canada\footnote{https://www.cbc.ca/news/canada/saskatchewan/privacy-commissioner-ehealth-ransomware-attack-1.5866119} that has affected millions of user files. Moreover, the cloud-based systems lack adequate responsiveness under network interruptions or high throughput scenarios, which can be devastating when in an emergency or monitoring fatal diseases. 

To this end, decentralized E-health analytics is a highly regarded paradigm, where each patient owns a personalized predictive model that is locally deployed (e.g., a smartphone application), allowing all the sensitive data to be retained on her/his personal device as all the analyses can be performed on-device. Such decentralization also benefits the service efficiency as it minimizes the dependency on network connectivity. Meanwhile, decentralized on-device E-health analytics brings new challenges in building an accurate predictive model without the need for collecting a large training dataset from multiple users. A naive solution is to train a personalized model on each user device individually; however, it can hardly produce satisfactory prediction accuracy given the limited amount of data on a single device \cite{bellet2018personalized}. 

Recently, federated learning (FL) \cite{konevcny2015federated, konevcny2016federated, mcmahan2017communication} has emerged an effective way of learning on-device models. Specifically, each device trains a predictive model on its local data with stochastic gradient descent (SGD). Then, the computed gradients from all devices are uploaded to the central server, which are then summarized and used to update the universal model shared across all devices \cite{konevcny2015federated}.  
However, FL is still essentially a centralized paradigm, as a server is indispensable for coordinating all participating devices and maintaining a global state of the system. This is different from the \textbf\textit{fully decentralized} setting with no central entities, which is the only available option in many applications (e.g., IoT and peer-to-peer networks) \cite{bellet2018personalized}. Hence, subsuming E-health analytics under the FL framework fails to exempt the demand on cloud-based infrastructures, and it is known to fall short in scalability \cite{bellet2018personalized} when handling a rapidly growing set of devices. Furthermore, while E-health analytics are highly personalized services, FL essentially forces all users to share a generic model in common, making the learned model biased towards the majority user group having dense data.

To this end, we investigate the challenging problem of allowing users' personal devices to collaborate with each other in fully decentralized E-health analytics, so as to optimize and personalize their purely local models. In this paper, we propose a novel solution to fully decentralized and personalized E-health analytics. As the core objective in this research problem, the decentralization and personalization of each on-device predictive model are established upon the block coordinate descent (BCD) optimization algorithm \cite{zhang2017convergence}. Compared with SGD, BCD is a gradient-free optimizer and can intrinsically avoid the gradient vanishing issue \cite{zhang2017convergent} caused by stacking deep layers. Hence, it is better suited to the fully decentralized learning setting where model updates are scarce, and as pointed out by \cite{lau2018proximal}, BCD fully supports distributed and parallel implementation. The basic idea of BCD is to find an exact solution of a subproblem with respect to a subset (i.e., block) of variables \cite{tseng2001convergence} rather than finding the solution for all variables simultaneously, so as to improve computational efficiency. However, while convergence analysis on BCD is feasible under the assumption that each on-device model is strongly convex \cite{bellet2018personalized}, the objective functions of DNN models are commonly non-convex. Despite the recent approach to global convergence of general DNNs via BCD \cite{zeng2019global}, the method focuses only on the centralized setting. Moreover, it requires each on-device dataset to follow a similar distribution, thus overlooking the personalized nature of E-health analytics.

To fill the gap, we design a \textbf{decentralized BCD (D-BCD)} framework for DNN-based E-health models. Specifically, by treating each layer in the DNN as a block in BCD, we further propose a novel perspective -- that is, each device can actually contain several blocks (i.e., layers) rather than just one block, while being compliant to the BCD paradigm. As such, each of the blocks in the entire neural network will be updated once in each iteration. Moreover, we build our D-BCD framework with decentralized multi-layer perceptrons (MLPs) as the backbone DNN models, where each device holds a unique model trained on its private dataset. To maximize the effectiveness of collaborative model learning, we further propose an innovative, similarity-based inter-device communication protocol by allowing each on-device model to learn from neighbors with high affinity to it. We also investigate how pairwise communication in D-BCD alleviates the common cold-start and data imbalance problem\cite{zhang2018discrete} in personalized E-health. Experimental results on real-world health data analytics demonstrate that D-BCD's prediction accuracy is on par with centralized peer methods, while offering numerous advantages in terms of decentralization, personalization, and significantly faster convergence. We summarize the major contributions of this paper as follows:
\begin{itemize}
\item In light of the arising privacy and robustness concerns on cloud-based/centralized E-health services\cite{zhang2021graph, wang2021fast}, we study a fairly new E-health paradigm that is fully decentralized and personalized. This paradigm allows each E-health user to possess a customized on-device predictive model, thus retaining their sensitive data on their personal devices and being able to receive instant medical advice. 
\item We propose a novel D-BCD framework to facilitate learning personalized deep models under the fully decentralized setting, where no central entities are needed to coordinate the joint learning process. In D-BCD, we implement decentralized on-device MLPs, which are optimized using our developed similarity-based collaborative learning protocol in a block-wise manner.  
\item We conduct extensive benchmarking experiments on real-world datasets, which demonstrates the effectiveness and efficiency of the D-BCD framework in addressing different E-health analysis tasks.
\end{itemize}

\section{Related Work}

Decentralized algorithms are sometimes referred to as gossip algorithms because the learned knowledge spreads along the edges specified by the communication graph \cite{1638541, XIAO200465, chen2021learning, wang2020next}. Its principal objective is to train models in a device network without a central coordinator (e.g. parameter server\cite{reisizadeh2020fedpaq}, \cite{bonawitz2019towards}) but instead only requiring on-device computation and local communication with neighboring devices (e.g. exchanging partitioned gradients or model updates)\cite{pappas2021ipls}. The optimization problem of decentralized device networks based on SGD algorithms has been thoroughly studied. \cite{jakovetic2014fast, scaman2018optimal, doan2020fast, uribe2020dual} developed optimal algorithms for deterministic (non-stochastic) convex objective functions. \cite{shamir2014distributed, rabbat2015multi} derived the rate for the stochastic setting assuming the distributions on all nodes are equal. There are also algorithms free of such i.i.d. assumption like \cite{koloskova2019decentralized, yu2019distributed, assran2019stochastic}. 

Unlike SGD (gradient-based method) that makes use of backpropagation to compute gradients of network parameters \cite{rumelhart1986learning}, gradient-free methods have been recently adapted to the DNN training. The representative algorithms are BCD \cite{carreira2014distributed, zhang2017convergence, askari2018lifted} and Alternating Direction Method of Multipliers (ADMM) \cite{taylor2016training}. Gradient-free algorithms perform well in dealing with non-differentiable nonlinearities and accordingly, they can potentially avoid the gradient vanishing issue \cite{zhang2016efficient}. Auxiliary coordinates method is a sort of mathematical device to design optimization algorithms that are suited for any specific nested architecture, which enables BCD algorithms to optimize DNN models \cite{carreira2014distributed}. The auxiliary coordinates method used in this paper is called three-splitting formulation, where all weight matrices in all hidden layers and all activation vectors are concatenated respectively into two separate blocks and updated together with the weight matrix of the output layer \cite{lau2018proximal}. Convergence results of such approach based on different activation functions are provided in \cite{xu2013block, xu2017globally, gu2020fenchel}.

Though works like \cite{bellet2018personalized} and \cite{vanhaesebrouck2017decentralized} applied BCD to improving local devices by communicating with neighbor devices that have similar objectives, the adopted loss function is convex and the validation experiments are based on naive tasks (e.g. linear classification and mean estimation). In actual telemedicine practice, we need more complex non-convex DNN models to tackle the intricate classification tasks based on data collected by sensors. Therefore, we propose to deploy BCD in a decentralized setting with non-convex loss functions. Meanwhile, the communication graphs in \cite{bellet2018personalized} and \cite{vanhaesebrouck2017decentralized} do not take both the task relatedness and the communication cost into consideration.  

\section{Methodology} \label{sec:method}
In this section, we introduce our technical pathways towards D-BCD for personalized E-health analytics. We first outline the baseline models deployed at the user side and the device network. Then, we present our approach to decentralized optimization via a three-splitting formulation of the objective function and similarity-based collaborative learning.

\subsection{Preliminary} \label{sec:Preliminaries}
\textbf{Definition 1: Personalized Model.} Assuming an E-health application involves $A$ users, and each user device possesses a personalized deep model $\Phi(\cdot)$ parameterized by $\Theta_a$ ($a\leq A$). The $a$-th user device also hosts $N_a$ locally collected private data instances $\{(\textbf{x}_{an},\textbf{y}_{an})\}_{n=1}^{N_a}$, where $\textbf{x}_{an}$ denotes the input features (e.g., a medical record) and $\textbf{y}_{an}$ denotes the corresponding ground truth (e.g., the label of diagnosed disease). Then, the learning objective of is to minimize the gap between $\Phi(\x_{an} ; \Theta_a)$ and $\mathbf{y}_{an}$ for all $n\leq N_a$, $a \leq A$. Without loss of generality, we consider a generic case where $\Phi(\cdot)$ is an $N$-layer feedforward multilayer perceptron omitting the bias term:
\begin{eqnarray}\label{eq:DNN-model}
 \Phi(\x ; \Theta)=\sigma_N(\W_N\sigma_{L-1}(\W_{L-1}\cdots\W_2\sigma_1(\W_1\x )).
\end{eqnarray}
where $\W_{\cdot}$ denotes the weight matrix in each layer, $\sigma_\cdot$ is the non-linear activation function, and $\Theta:=\{\W_i\}_{i=1}^L$. The input and output layers are regarded as the $0$-th and the $L$-th layers, respectively. 

\textbf{Definition 2: Device Network.} The user devices can communicate with each other in the fully decentralized setting. Let $\mathcal{G} = (\mathcal{A}, \mathcal{E}, \mathcal{C})$ be a weighted graph, where $\mathcal{A} = \{1, \cdots ,A\}$ is the set of user devices (represented via indexes), and $\mathcal{E} \in \mathcal{A} \times \mathcal{A}$ is the set of edges between devices. $C \in \mathbb{R}^{A \times A}$ is a nonnegative weight matrix, where each entry $c_{i j} \in C$ indicates the weight of edge $(i, j) \in E$, which represents the communication cost between device $i$ and device $j$. $c_{a b} = 0 $ if $(a, b) \notin E$ or $a = b$. 
We use $\mathcal{G}_a$ to denote the set of $M$ neighbors of device $a$, where $b\in \mathcal{G}_a$ are top-$M$ devices with the lowest communication costs to $a$. 

\textbf{Problem Definition: Personalized On-Device E-health Analytics.} For all models in the device network $\mathcal{G}$, our goal is to learn the optimal $\overline{\Theta}:=\{\Theta_a\}_{a=1}^A$ that minimizes the following objective:
\begin{eqnarray}\label{eq:decen-empi-risk}
\mathcal{L} = \sum_{a=1}^A\frac{1}{N_a}\sum_{j=1}^{N_a}\ell(\Phi(\x_{aj};\Theta_a),\mathbf{y}_{aj}),
\end{eqnarray}
where $\ell$ is the loss function that quantifies the prediction error.

\subsection{Block Coordinate Descent with Variable Splitting}
Before presenting the details of the fully decentralized D-BCD, we first briefly introduce the BCD algorithm originally proposed in the centralized setting \cite{zeng2019global}. We start with a na\"ive situation where all devices are trained independently. The empirical loss minimization of a single on-device model in $\mathcal{G}$ can be expressed as:
\begin{eqnarray}\label{eq:empi-risk}
&\mathop{\min}\limits_{\Theta_a} \frac{1}{N_a}\sum_{j=1}^{N_a}\ell(\Phi(\x_j;\Theta_a),\mathbf{y}_j).
\end{eqnarray}
As the variables are coupled via the deep neural network architecture in Eq.~\ref{eq:DNN-model}, the problem is highly non-convex and computationally intractable. In this regards, variable splitting is a well-acknowledged trick \cite{zeng2019global,taylor2016training,lau2018proximal} that generates some auxiliary variables to decouple the original variables, which makes the problem more tractable. Essentially, the $L$-layer network structure in Eq.~\ref{eq:empi-risk} can be naturally formulated as the following two-splitting form:
\begin{align}\label{eq:two-splitting}
&\mathop{\min}\limits_{\Theta, \mathcal{V}} \frac{1}{N_a}\sum_{j=1}^{N_a}\ell(\mathbf{v}_{Lj},\mathbf{y}_j) + \sum_{i=1}^{L}r_i(\mathbf{W}_i) + \sum_{i=1}^{L}s_i(\mathbf{v}_i), \nonumber\\
&\textrm{s.t. }  \  \mathbf{v}_i=\sigma_i(\mathbf{W}_i\mathbf{v}_{i-1}),\ i=1, \dots ,L,
\end{align}
where $\mathcal{V}:= \{\mathbf{v}_i\}_{i=1}^L$, $(\mathbf{v})_{:j}$ is the $j$-th column of $\mathbf{v}_N$, while $r_i$ and $s_i$ are non-negative functions that reveal the prior of the weight matrix $\mathbf{W}_i$ and state vector $\mathbf{v}_i$ respectively at each layer. Following \cite{zeng2019global}, we set both $r_i$ and $s_i$ to the 
$L2$ regularization function.

However, in the $i$-th constraint of the two-splitting form of a model's objective (i.e., Eq.~\ref{eq:two-splitting}), the entanglement between $\mathbf{W}_i$ and $\mathbf{v}_{i-1}$ imposed by nonlinearity $\sigma_i(\cdot)$ can bring difficulties in efficiently solving it \cite{zeng2019global}. Therefore, another set of auxiliary variables $\mathcal{U}:= \{\mathbf{u}_i\}_{i=1}^L$ is introduced in BCD, where $\mathbf{u}_i = \mathbf{W}_i\mathbf{v}_{i-1}$, i.e., the state vector prior to the nonlinear activation. The modified form of Eq.~\ref{eq:two-splitting} is called three-splitting formulation \cite{taylor2016training}. 

In a nutshell, the BCD method for an independent device is a backward, cyclical variable update process. Specifically, the variables are updated from the output (i.e, $L$-th) layer to the input (i.e., $0$-th) layer. In each layer $i$, each of the three blocks $\mathbf{W}_i, \mathbf{u}_i$ and $\mathbf{v}_i$ will be updated cyclically while fixing the other two at each time. 

\subsection{Multi-Device Block Coordinate Descent}
We hereby extend the plain BCD method to the decentralized environment where multiple on-device models are trained asynchronously on distributed E-health data. Let $\overline{\mathcal{Z}}:=\{(\x_{aj}, \mathbf{y}_{aj})\}_{j=1, a=1}^{N_a,A}$ be the set of all training samples across the device network. Then, the three-splitting formulation for all model parameters $\overline{\Theta}$ w.r.t. Eq.~\ref{eq:decen-empi-risk} can be written as: 
\begin{align}\label{eq:multi_3s}
&\mathcal{L}_0(\overline{\Theta}, \overline{\mathcal{V}}):=\sum_{a=1}^A\mathcal{L}_{N_a}(\{(\textbf{x}_{an},\textbf{y}_{an})\}_{n=1}^{N_a})+\sum_{a=1}^A\sum_{i=1}^Lr_i({\mathbf{W}_{ai}})\nonumber \\
&+\sum_{a=1}^A\sum_{i=1}^Ls_i({\mathbf{v}_{ai}}),  \\
 &\textrm{s.t. } \mathbf{u}_{ai} = \mathbf{W}_{ai}\mathbf{v}_{a(i-1)}, \  \mathbf{v}_{ai}=\sigma_i(\mathbf{u}_{ai}),\ i=1, \dots ,L, \nonumber
\end{align}
where $\mathcal{L}_{N_a}(\{(\textbf{x}_{an},\textbf{y}_{an})\}_{n=1}^{N_a}):= \frac{1}{N_a}\sum_{j=1}^{N_a}\ell(\mathbf{v}_{aLj},\mathbf{y}_{aj})$, $\overline{\mathcal{V}}:= \{\mathbf{v}_{ai}\}_{i=1, a = 1}^{L,A}$ and $\overline{\mathcal{U}}:= \{\mathbf{u}_{ai}\}_{i=1,a=1}^{L,A}$. We address the following alternative unconstrained problem to solve Eq.(\ref{eq:multi_3s}):
\begin{align}\label{eq:obj}
&\mathop{\min}\limits_{\overline{\Theta},\overline{\mathcal{V}},\overline{\mathcal{U}}}\mathcal{L}(\overline{\Theta},\overline{\mathcal{V}},\overline{\mathcal{U}}):=\mathcal{L}_0(\overline{\Theta}, \overline{\mathcal{V}})+\sum_{a=1}^A\sum_{i=1}^L(\frac{\gamma}{2}\parallel\mathbf{v}_{ai}-\sigma_i(\mathbf{u}_{ai})\parallel_F^2 \nonumber \\ 
&+\frac{\alpha}{2}\parallel\mathbf{u}_{ai}-\mathbf{W}_{ai}\mathbf{v}_{a(i-1)}\parallel_F^2),
\end{align}
where $\gamma, \alpha > 0$ are two hyperparameters. Compared with the centralized BCD, the key difference of D-BCD is that, models in D-BCD are personalized rather than unified for each E-health service user, offering more flexibility to enable different data distributions across the network. In the following subsections, we will provide details on how we design D-BCD for optimizing Eq.(\ref{eq:obj}). 

\subsection{Similarity-based Collaborative Learning} \label{sec:inter-device}
However, a non-trivial obstacle needs to be overcome before the E-health service users can enjoy the benefit of personalization. Specifically, the fully decentralized paradigm leads to the fact that each on-device model heavily relies on the local data about the user to facilitate personalization. Given the limited amount of data on each device, the performance of the resulted personalized models will be suboptimal. Unlike the centralized learning paradigms where a central server can leverage knowledge from all participating devices (e.g., uploaded gradients in FL) to learn a performant but non-personalized global model, D-BCD does not allow such resource-intensive training process due to the absence of the central server.

Hence, we design a similarity-based inter-device communication protocol to support training. Instead of having a central authority that manages all devices, D-BCD can only allow devices to communicate with each other to exchange the knowledge. Due to the limited communication bandwidth in real-world E-health applications, for each device, such inter-device communication is restricted to only a small group of devices based on the predefined communication cost $c$ in the Section \ref{sec:Preliminaries}. On top of that, we propose to intervene the learning process with user-wise similarity. Let $h(a, b)$ be the similarity between device $a$ and its  neighbor devices $b\in \mathcal{G}_a$, and let $H_a^M= \sum_{b\in \mathcal{G}_a} h(a, b)$. Then, the similarity-based model update for the $k$-th training iteration is:
\begin{align}\label{eq:aggregation}
\Theta_{a}^{k} \leftarrow (1-\mu)\Theta_{a}^k + \frac{\mu}{H_a^M} \sum_{b\in \mathcal{G}_a} h(a, b)\Theta_b^{k},
\end{align}
where $\mu$ is a non-negative trade-off parameter and $\Theta_{b}^{k}$ denotes the neighbor models' parameters. As D-BCD supports asynchronous model update, we use $\Theta_{b}^{k}$ to denote the newest possible model parameters of every $b \in \mathcal{G}_a$ for notation simplicity.
The rationale of our similarity-based model aggregation is that, D-BCD with inter-device communication can improve the model performance on sparse data. Meanwhile, the choice of $h(\cdot)$ is versatile, and such collaborative learning essentially helps each local model to incorporate more knowledge from relevant neighbor models, thereby guaranteeing the performance of each personalized model.
Additionally, all $\Theta_{b}^{k}(b\in \mathcal{G}_a)$ are masked by the parameters of $b$'s neighbors before sharing, which hinders the possible inference attack from $a$ since it lacks the information of $\mathcal{G}_b$.

\subsection{Learning Personalized E-health Models with D-BCD}
With the block coordinate descent principle, we optimize one layer of the model on each device at one time. Let $d_i$ be the number of hidden dimension for the $i$-th layer of the on-device feedforward network, the optimization process of our D-BCD algorithm are described in Algorithm \ref{Ag:Decen-3-BCD}.

\begin{algorithm}[]
\LinesNotNumbered
\SetAlgoLined
Set $k \leftarrow 1$\;
\For{$a=1,\cdots,A$}{
	\For{$i=1,\cdots,L$}{
		choose $\mathbf{W}_{ai}^0 \in \mathbb{R}^{d_i \times d_{i-1}}, \mathbf{v}_{ai}^0 \in \mathbb{R}^{d_i}$ and $ \mathbf{u}_{ai}^0 \in \mathbb{R}^{d_i}$\;
	}
}
\Repeat{termination condition satisfied}{
	\For{$a=1,\dots,A$}{
		$\mathbf{v}_{a0}^{k} \leftarrow \mathbf{x}_a $\;
		$\mathbf{v}_{aL}^k \leftarrow \arg \min_{\mathbf{v}_{aL}}\{\mathbf{s}_N(\mathbf{v}_{aL})+\mathcal{R}_{m_a}({\mathbf{v}_{aL}};\mathbf{Y}_a)+ \frac{\gamma}{2} {\parallel {\mathbf{v}_{aL}}}-{\mathbf{u}_{aL}^{k-1}} \parallel_F^2 + {\frac{\alpha}{2}\parallel {\mathbf{v}_{aL}}-{\mathbf{v}_{aL}^{k-1}} \parallel_F^2} \}$\;
		$\mathbf{u}_{aL}^k \leftarrow  \arg \min_{\mathbf{u}_{aL}}\{\frac{\gamma}{2} {\parallel {\mathbf{v}_{aL}^k}-\mathbf{u}_{aL} \parallel_F^2} + {\frac{\gamma}{2}\parallel\mathbf{u}_{aL}}-\mathbf{W}_{aL}^{k-1}\mathbf{v}_{a(L-1)}^{k-1} \parallel_F^2 \}$\;
		$\mathbf{W}_{aL}^k \leftarrow  \arg \min_{\mathbf{W}_{aL}}\{\mathbf{r}_N(\mathbf{W}_{aL})+ \frac{\gamma}{2} {\parallel \mathbf{u}_{aL}^k}-\mathbf{W}_{aL}\mathbf{v}_{a(L-1)}^{k-1} \parallel_F^2 + {\frac{\alpha}{2}\parallel {\mathbf{W}_{aL}}-{\mathbf{W}_{aL}^{k-1}} \parallel_F^2} \}$\;
		\For{$i=L-1,\dots , 1$}{
		$\mathbf{v}_{ai}^k \leftarrow  \arg \min_{\mathbf{v}_{ai}}\{\mathbf{s}_i(\mathbf{v}_{ai})+ \frac{\gamma}{2} {\parallel {\mathbf{v}_{ai}}}-\sigma_i(\mathbf{u}_{ai}^{k-1}) \parallel_F^2 + {\frac{\alpha}{2}\parallel {\mathbf{u}_{a(i+1)}^k}-\mathbf{W}_{a(i+1)}^{k}\mathbf{v}_{ai} \parallel_F^2} \}$\;
		$\mathbf{u}_{ai}^k \leftarrow  \arg \min_{\mathbf{u}_{ai}}\{\frac{\gamma}{2} {\parallel {\mathbf{v}_{ai}^k}-\sigma_i(\mathbf{u}_{ai}) \parallel_F^2} + {\frac{\gamma}{2}\parallel\mathbf{u}_{ai}}-\mathbf{W}_{ai}^{k-1}\mathbf{v}_{a(i-1)}^{k-1} \parallel_F^2 + {\frac{\alpha}{2}\parallel\mathbf{u}_{ai}}-\mathbf{u}_{ai}^{k-1}\parallel_F^2\}$\;
		$\mathbf{W}_{ai}^k \leftarrow  \arg \min_{\mathbf{W}_{ai}}\{\mathbf{r}_i(\mathbf{W}_{ai})+ \frac{\gamma}{2} {\parallel \mathbf{u}_{ai}^k}-\mathbf{W}_{ai}\mathbf{v}_{a(i-1)}^{k-1} \parallel_F^2 + {\frac{\alpha}{2}\parallel {\mathbf{W}_{ai}}-{\mathbf{W}_{ai}^{k-1}} \parallel_F^2} \}$\;
		}
		$\Theta_{a}^k \leftarrow (1-\mu)\Theta_{a}^k + \frac{\mu}{H_a^M} \sum_{m=1}^M h(a, m)\Theta_{g(a,m)}^{k-1}$\;
	}
	$k \leftarrow k+1$\;
}
\caption{D-BCD with Similarity-based Collaborative Learning}
\label{Ag:Decen-3-BCD}
\end{algorithm}

Specifically, in Algorithm \ref{Ag:Decen-3-BCD}, we first initialize the parameters on all devices via lines $1$ - $7$ when the training starts. For each device, lines $10$ - $13$ feed the input into the on-device model, thus performing a local BCD update for the $L$-th (bottom) layer the model. Analogously, we then update all on-device models' parameters from the $L-1$-th layer to the first layer via lines $14$ - $16$. Finally, we leverage our proposed similarity-based collaborative learning in line $18$ to enhance each local model by selectively aggregating the knowledge from highly relevant neighbors.

\section{Experiment}  \label{sec:experiment}
In this section, we conduct experiments on three real-world datasets to verify the effectiveness of D-BCD in different E-health analytic tasks.

\subsection{Datasets} \label{sec:datasets}
We use three experimental datasets in our experiments, which are introduced below.

Firstly, we use the \textbf{Sleep Cassette (SC)} dataset \cite{mourtazaev1995age} that contains 153 whole-night polysomnography (PSG) recordings and sleep stage markers are introduced to examine the proposed algorithm. Electroencephalogram (EEG) based sleep stage classification is vital to the diagnosis of diseases like diabetes, hypomnesia and excessive daytime sleepiness \cite{skeldon2014mathematical}. We cascade each 30-second two-channel (PFz-Cz and Pz-Oz) $100$Hz EEG signals in SC to 
generate the model input. SC also provides Rechtschaffen and Kales (R\&K) sleep stage \cite{rechtschaffen1968manual} annotations, which are mapped into three discrete labels representing awake, non-rapid eye movement sleep (NREM), and rapid eye movement sleep (REM), respectively. We use the ratio of $60\%$, $20\%$, and $20\%$ for partitioning the training set, validation set, and test set.

Secondly, we train the proposed method to detect Obstructive Sleep Apnea (OSA), a common sleep-related breathing disorder\cite{ye2021fenet,espiritu2021health}. The PhysioNet apnea-ECG dataset \cite{ichimaru1999development} collected $70$ overnight ECG recordings and apnea event labels from different patients. We follow algorithm \cite{cai2020qrs} to convert the ECG signals into pulse signals, for which we use a $60$-dimensional vector to represent the RR-interval (time interval between two R-peaks) series at every minute, and a binary label to indicate whether there is an apnea event in this period for the given patient. The resulted dataset is called \textbf{P-Pulse}. Each recording is regarded as one local dataset of a device, i.e., $A=70$ in P-Pulse. We follow the same partitioning ratio as used in SC.

In addition to the two E-health datasets, we further conduct experiments on \textbf{MNIST} \cite{lecun1998gradient}, a dataset that contains images and labels of $70000$ handwritten digits. We select MNIST as it is a widely used dataset for benchmarking decentralized machine learning algorithms. Following the settings in \cite{bistritz2020distributed}, samples in MNIST are randomly and evenly distributed across  $50$ devices, each of which contains $1400$ samples. Data split for each device also follows the same ratio as in SC and P-Pulse.

\subsection{Experimental Setting}
As demonstrated in the Section \ref{sec:method}, the DNN model deployed in each device is simply an $L$-layer feedforward network. To further showcase the generalization capability of the proposed framework, we specify $r_i = s_i = 0$, $d_i = d \ (i = 1, \cdots\ L-1)$, $\ell$ as the cross-entropy loss and $\sigma$ as the ReLU activation function for all layers. We compare D-BCD with the following baseline learning algorithms:
\begin{itemize}
\item \textbf{C-SGD}: The central server stores all the data and trains a global model via SGD algorithm.
\item \textbf{D-SGD}: Each device has a unique model and uses SGD to optimize model parameters. Inter-device communication is allowed.
\item \textbf{C-BCD}: The central server stores all the data and trains a global model via BCD algorithm.
\item \textbf{I-BCD}: Each device has a unique model and uses BCD to optimize model parameters without inter-device communications.
\end{itemize}

All optimal hyperparameters are determined via grid search. For the number of neighbors in the device network, it is searched in $M = \{0, 5, 10, 50\}$. The searching space of $L$ and $d$ are $\{4,8,16,32,64\}$ and $\{32, 64, 128, 256\}$, respectively. The trade-off factor $\mu$ is tuned in $\{0.01, 0.1, 0.5, 0.9\}$ and $\gamma$ and $\alpha$ are tuned in $\{0.1, 0.5, 1, 5, 10\}$. For D-SGD algorithm, the batch size is $128$ and the learning rate is $0.05$. The model is trained on High Performance Computing (HPC) Systems with $6$ NVIDIA V100 GPUs.

In general, the communication cost $c$ depends on physical limitations, like the distance between devices while transferring data via wireless methods. 
For the E-healthcare scenario, the model parameters are typically shared via the internet to reduce communication latency, which levels the $c$ between different devices. In this case, 
three randomly generated undirected weighted graphs with the maximum node degree of $50$ are introduced to simulate the communication condition of all devices. For MNIST, we defined the similarity between any two devices as $1$, since they are subject to identical distribution. For SC and P-Pulse, the similarity $h(a,b)$ between devices $a$ and $b$ is calculated via:
\begin{equation}
\begin{split}
h(a,b) = \frac{\mathbf{e}_a \cdot \mathbf{e}_b}{||\mathbf{e}_a|| \ ||\mathbf{e}_b||},
\end{split}
\end{equation}
where $\mathbf{e}_{\cdot}$ is the vector representing the user-specific information. For example, the elements in $\mathbf{e}_a$ are normalized age, sex and body mass index (BMI) information. 
We use accuracy (Acc), macro-precision (mPre) and macro-recall (mRec) to evaluate the performance of our approach and four baselines on multi-class classification task (MNIST and SC), and use accuracy (Acc), precision (Pre) and recall (Rec) for binary classification task (P-Pulse). 


\subsection{Effectiveness Evaluation} \label{sec:Effectiveness}

\textbf{Overall Performance.} 
We report the performance of D-BCD on three datasets in Table \ref{tab:effectiveness}. The best performance of D-BCD on MNIST is reached when $M=50$, $d=128$, $L=4$, $\mu=0.01$, $\alpha = \gamma = 1$. The optimal setting for SC is $M=0$, $d=256$, $L=64$, $\mu=0.01$, $\alpha = \gamma = 1$, The optimal setting for P-Pulse is the same as SC except $L=8$. In general, centralized methods (C-SGD and C-BCD) perform better on MNIST while decentralized methods (D-SGD and D-BCD) perform better on P-Pulse.
One of the most prominent points is that I-BCD lags behind on MNIST but outperforms all other methods on P-Pulse in terms of accuracy and recall, which is likely a result of several factors. On the one hand, the local data of each device in P-Pulse is sufficient for training independent models. On the other hand, when the data distribution has high diversity among the neighbor devices, it may result in excessive noise in the model aggregation step. We have conducted additional experiments in the Section \ref{sec:Simulation} as a further investigation on this observation.
It is also noticeable that D-BCD always stands between C-BCD and I-BCD, which is in line with our intuition that inter-device communication will bring significant performance gain upon individual learning schemes while closing the gap towards fully centralized ones. Another important observation is, D-BCD is comparable to D-SGD with performance difference less than $2$\% on both MNIST and P-Pulse, thereby highlighting the applicability of D-BCD for fully decentralized E-health applications.

\begin{table}[h]
\caption{Performance of D-BCD and baselines on MNIST and P-Pulse.}
\vspace{-0.5cm}
\begin{center}
\begin{tabular}{c|c|c|c|c|c|c}
\hline
Dataset & Metric & C-SGD & D-SGD & C-BCD & I-BCD & D-BCD \\
\hline
\multirow{3}{*}{MNIST}
& Acc & \textbf{0.9810} & 0.9482 & 0.9723 & 0.8952 & 0.9337  \\  
& mPre & \textbf{0.9817} & 0.9505 & 0.9732 & 0.9102 & 0.9357  \\
& mRec & \textbf{0.9813} & 0.9481 & 0.9725 & 0.8964 & 0.9334  \\
\hline
\multirow{3}{*}{SC}
& Acc & 0.8230 & \textbf{0.8410} & 0.7809 & 0.8407 & 0.8344 \\  
& mPre & 0.7331 & 0.7663 & 0.6686 & \textbf{0.7682} & 0.7611  \\
& mRec & 0.7858 & \textbf{0.8141} & 0.7264 & 0.8078 & 0.7975  \\
\hline
\multirow{3}{*}{P-Pulse}
& Acc & 0.9095 & 0.9120 & 0.8969 & \textbf{0.9139} & 0.9054 \\
& Pre & 0.8793 & 0.8809 & 0.8651 & \textbf{0.8828} & 0.8823 \\
& Rec & 0.8747 & 0.8804 & 0.8536 & \textbf{0.8839} & 0.8579 \\
\hline
\end{tabular}
\label{tab:effectiveness}
\end{center}
\vspace{0cm}
\end{table}

\begin{figure}[t]
\subfigure[]{
\hspace{-0.5cm}\includegraphics[scale=0.5]{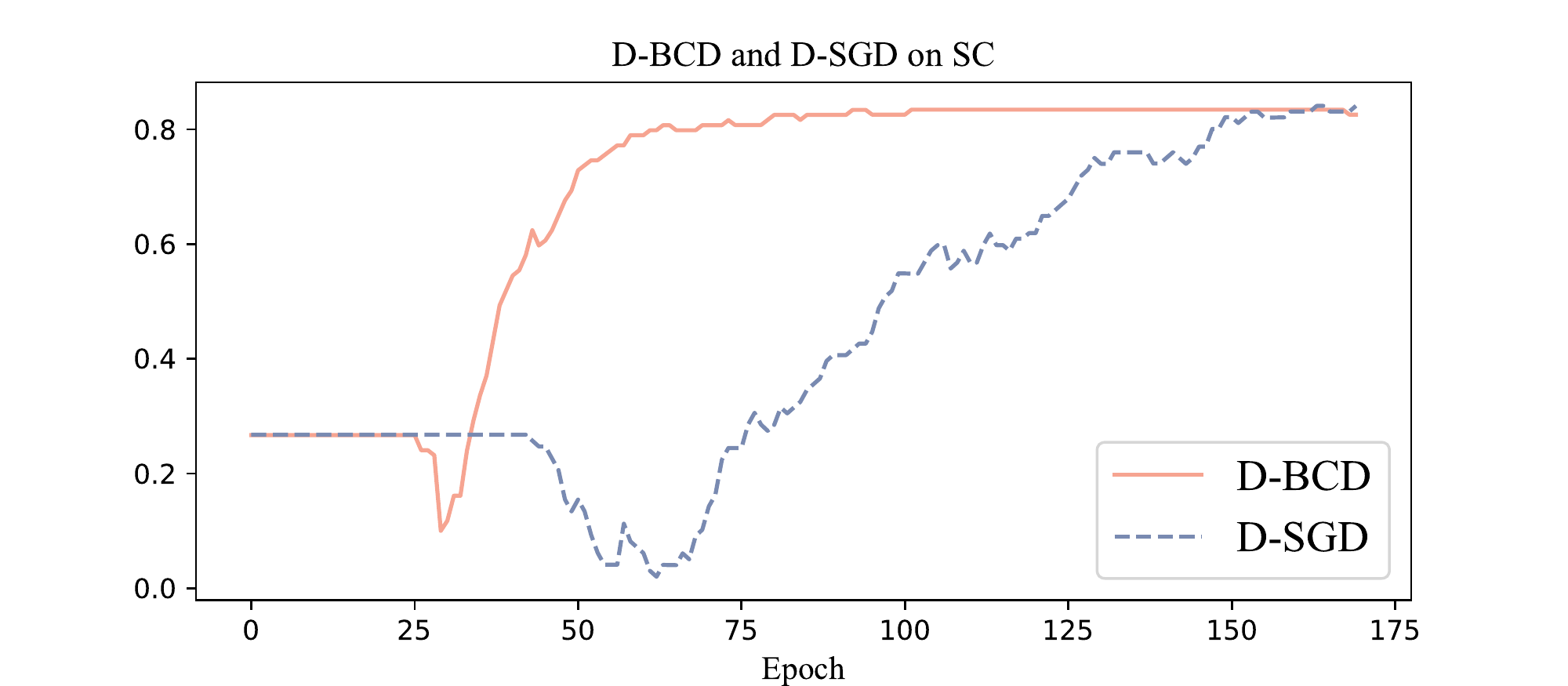}
}
\quad 
\subfigure[]{
\hspace{-0.5cm}\includegraphics[scale=0.5]{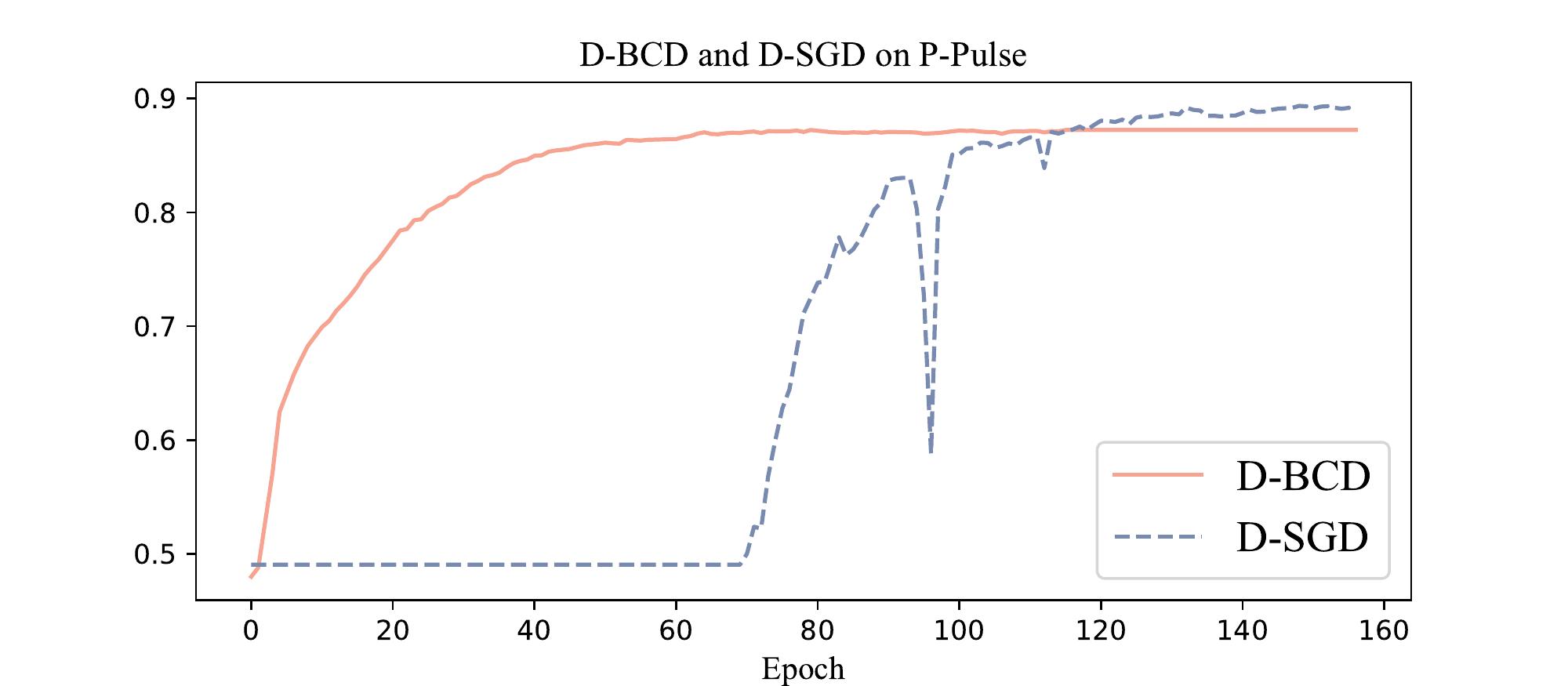}
}
\quad
\subfigure[]{
\hspace{-0.5cm}\includegraphics[scale=0.5]{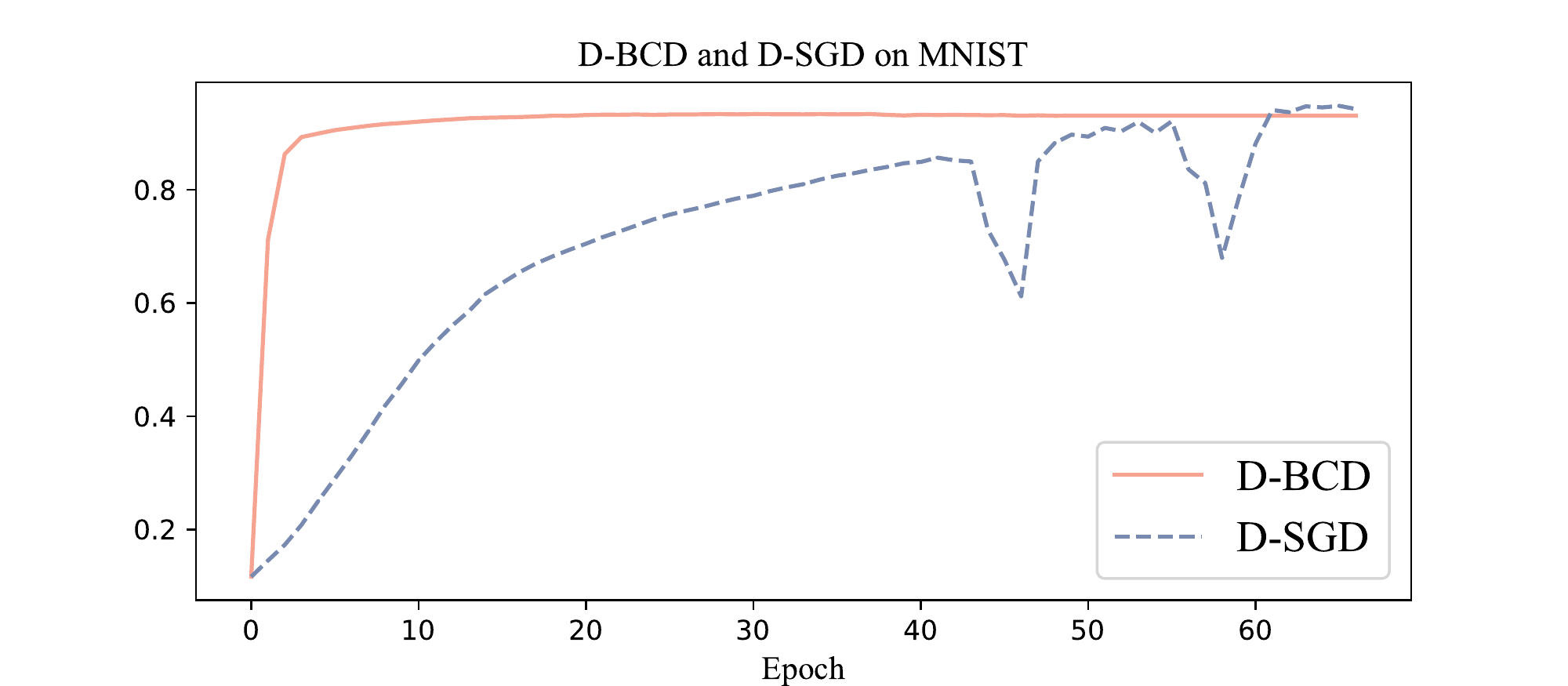}
}
\quad

\vspace{0cm}
\caption{(a), (b) and (c) are the test accuracy curves of the D-SGD and D-BCD on  SC, P-Pulse and MNIST, respectively.}
\label{fig:convergence}
\vspace{0cm}
\end{figure}

\textbf{Convergence Rate.} 
To showcase the suitability of D-BCD for on-device E-health analytics, we provide the accuracy-epoch curves of D-SGD and D-BCD in the training process in Fig.\ref{fig:convergence} to illustrate the superiority of D-BCD in terms of convergence rate. Apparently, D-BCD converges much faster than D-SGD, and is able to achieve high accuracy with substantially fewer training epochs. Notably, on SC and P-Pulse datasets, D-BCD reduces $36$\% and $38$\% of training epochs needed before convergence. In on-device E-health applications, the high training efficiency of D-BCD means longer battery life as it reduces the amount of energy-consuming communication and computation required. 

\begin{figure}[t]
\centering
\subfigure[]{
\begin{minipage}[t]{0.5\linewidth}
\includegraphics[scale=0.51]{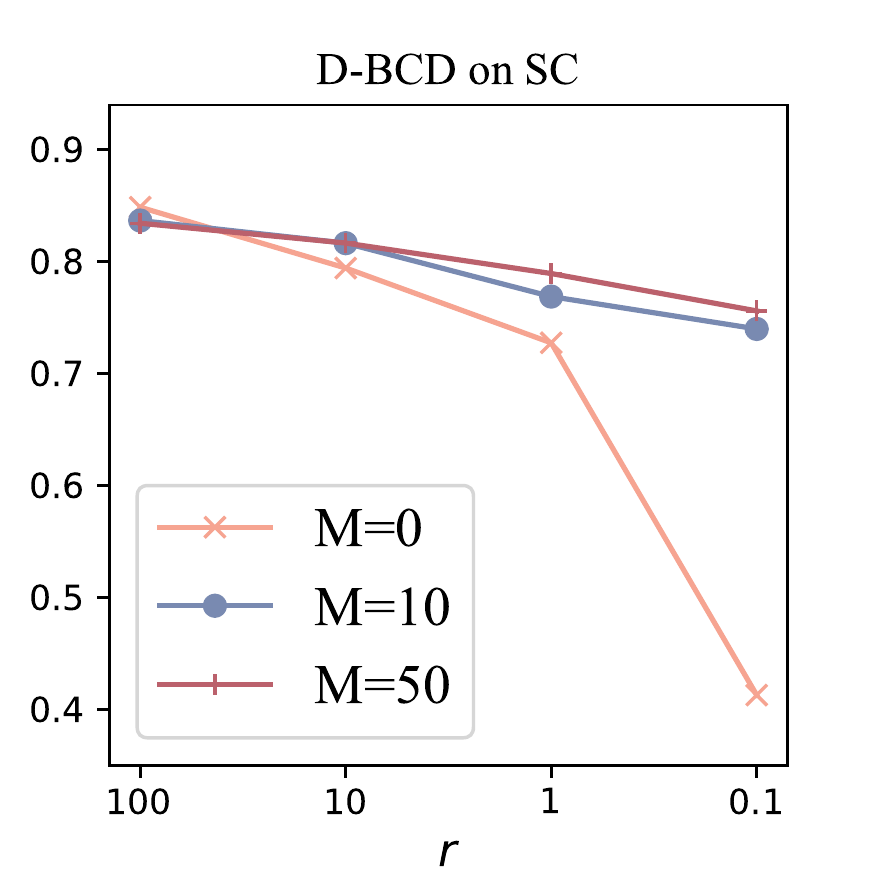}
\end{minipage}%
\label{fig:fig_sparsity2}
}%
\subfigure[]{
\begin{minipage}[t]{0.5\linewidth}
\includegraphics[scale=0.51]{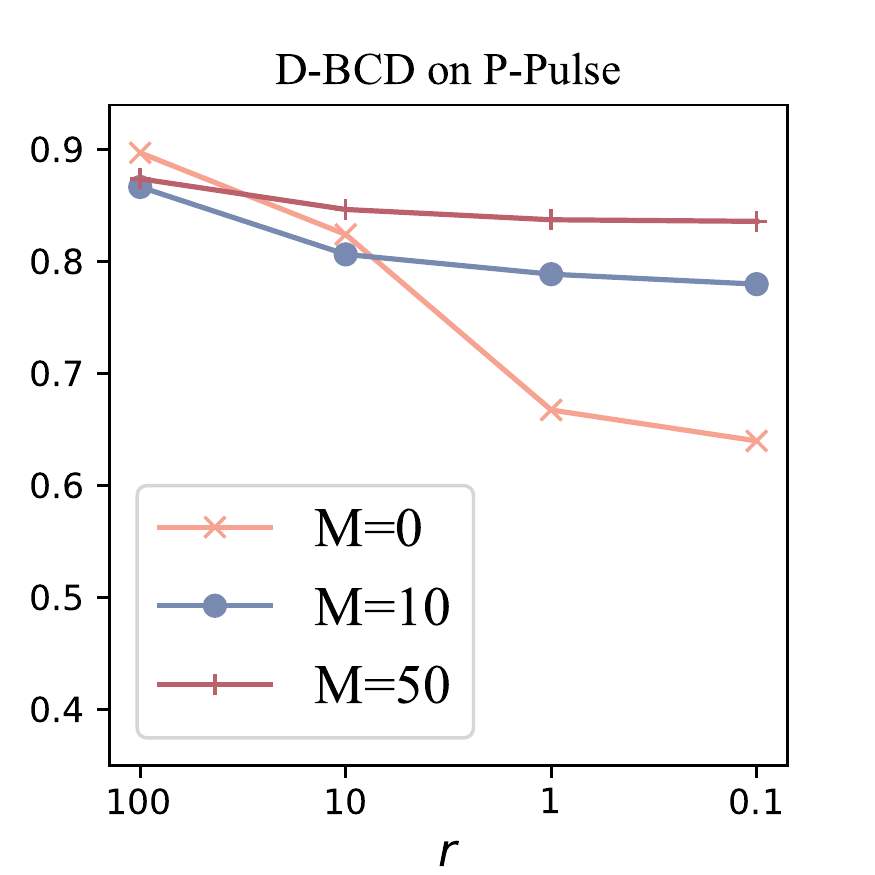}
\end{minipage}%
\label{fig:fig_sparsity1}
}%
\centering
\vspace{0cm}
\caption{(a) and (b) reflect the fluctuation of the performance of D-BCD w.r.t different data sparsity $r$ on SC and P-Pulse, respectively.}
\label{fig:Sparsity}
\vspace{0cm}
\end{figure}

\textbf{Handling Data Sparsity.}
We then test the effectiveness of the similarity-based collaborative learning for handing sparse on-device data. In SC and P-Pulse, we use $r$\% of the training set for each user and record the new performance resulted in Figure \ref{fig:Sparsity}. Furthermore, D-BCD with different number of neighbors ($M = 0 , 10 ,50$) is also tested w.r.t. different values of $r$. Note that D-BCD with $M=0$ is equivalent to I-BCD algorithm. As shown in Fig. \ref{fig:Sparsity}, all algorithms suffer from a distinct performance drop when $r$ decreases from $100$ to $0.1$. Intuitively, when $r$ goes lower, the available private data volume fails to effectively train a DNN model independently. As such, a dramatic performance drop can be observed when there is no help from neighbor devices ($M=0$). The accuracy of D-BCD with $10$ or $50$ neighbors, on the contrary, is stable even if only $0.1$\% training data is available. This experiment demonstrates the main advantage of decentralized architecture and suggests that our proposed similarity-based collaborative learning can mitigate the data sparsity issue on personal devices.

\begin{figure}[]
\centering
\subfigure[]{
\begin{minipage}[t]{0.5\linewidth}
\includegraphics[scale=0.52]{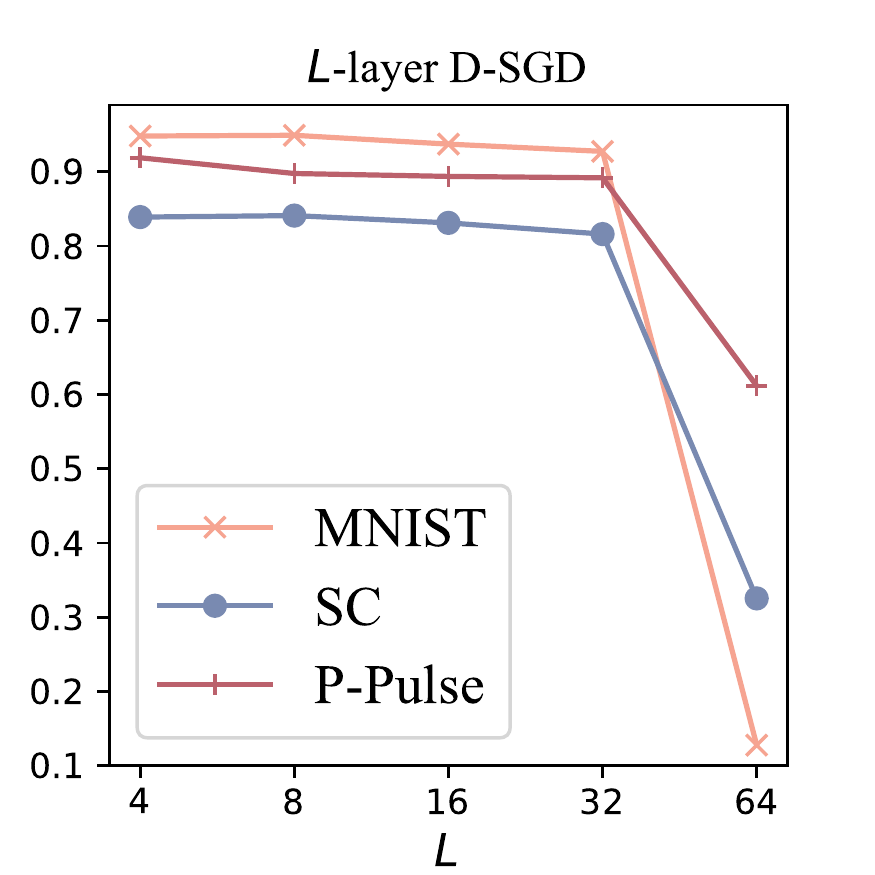}
\end{minipage}%
\label{fig:hyper_L1}
}%
\subfigure[]{
\begin{minipage}[t]{0.5\linewidth}
\includegraphics[scale=0.52]{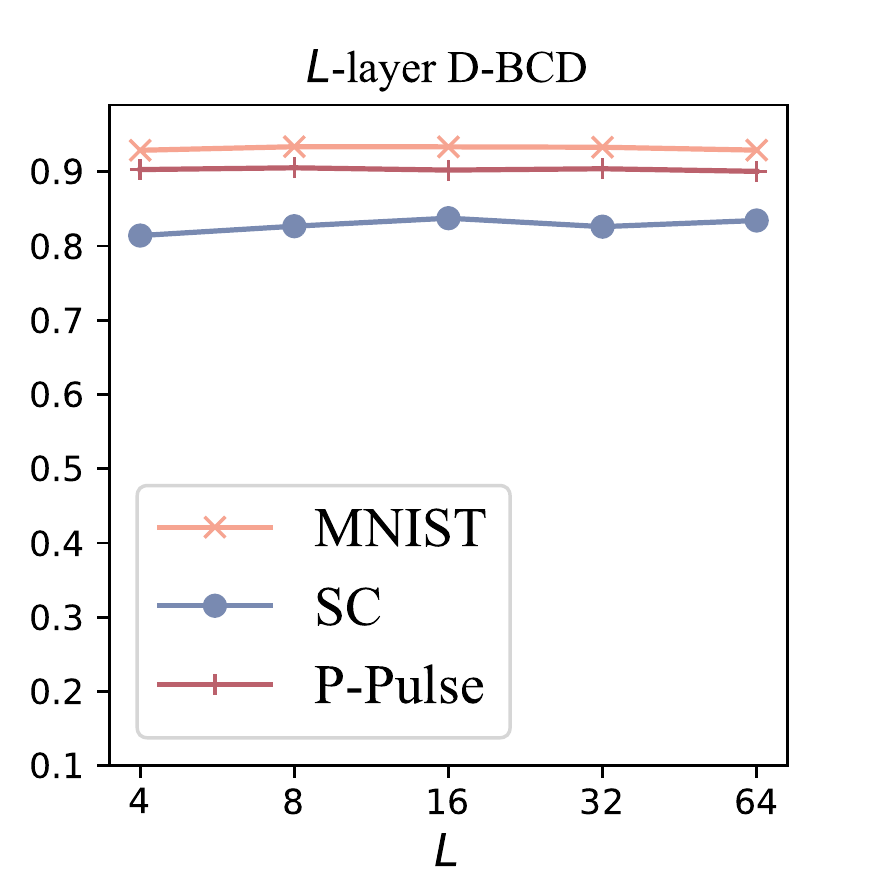}
\end{minipage}%
\label{fig:hyper_L2}
}%
\centering
\vspace{-0.05cm}
\caption{The test accuracy of D-SGD (a) and D-BCD (b) with different number of layer $L$ on all three datasets.}
\label{fig:hyper_layer}
\vspace{0cm}
\end{figure}

\subsection{Impact of Hyperparameters}  \label{sec:Hyperparameters}
We examine the effect of three key hyperparameters, namely $L$, $M$, and $\mu$ on the performance of D-BCD. First, we test the accuracy of D-BCD with different network depths $L$ with both D-SGD and D-BCD, where the results are shown in Fig.~\ref{fig:hyper_layer}. All other hyperparameters follow the optimal setting in this study. The results of D-SGD on all three databases (see Fig.~\ref{fig:hyper_L1}) indicate that D-SGD perform well when $L<32$ but is unable to train a 64-hidden-layer MLPs, which is typically due to the gradient vanishing phenomenon. However, D-BCD (see Fig.~\ref{fig:hyper_L2}) is much less insensitive to $L$ as the performance remains stable when $L$ varies. Obviously, BCD maintains its capability of mitigating gradient vanishing issues when transferring from the centralized settings\footnote{https://github.com/timlautk/BCD-for-DNNs-PyTorch} to fully decentralized environments. 

\begin{figure}[]
\centering
\subfigure[]{
\begin{minipage}[t]{0.5\linewidth}
\includegraphics[scale=0.52]{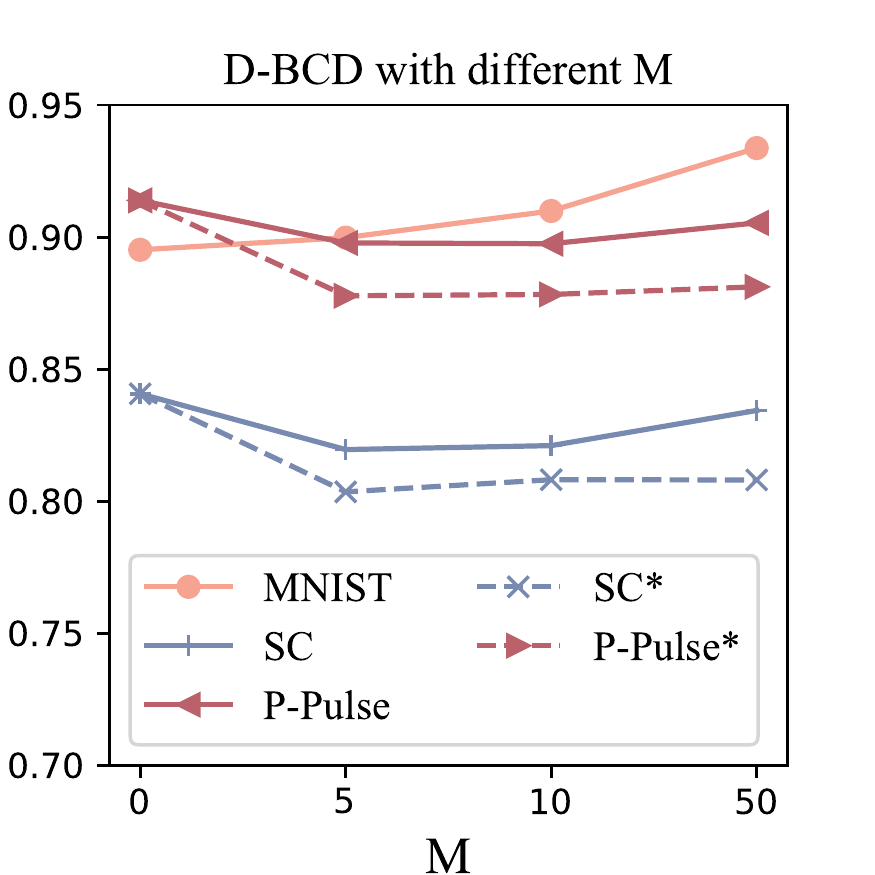}
\end{minipage}%
\label{fig:hyper_M}
}%
\subfigure[]{
\begin{minipage}[t]{0.5\linewidth}
\includegraphics[scale=0.52]{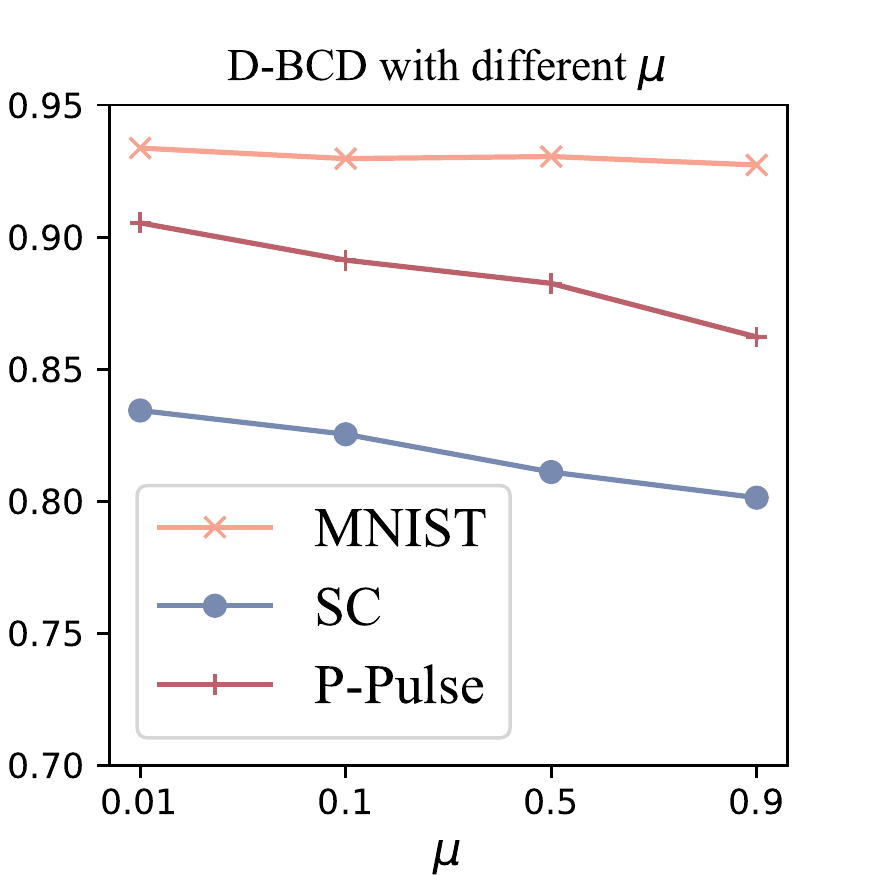}
\end{minipage}%
\label{fig:hyper_mu}
}%
\centering
\vspace{0cm}
\caption{The impact of hyperparameters of D-BCD on the test accuracy. (a) and (b) reflect the fluctuation of test accuracy in response to the varying $M$ and $\mu$, respectively.}
\label{fig:hyper_communicaion}
\vspace{-0.25cm}
\end{figure}

Next, we fix $\mu=0.01$ and enlarge $M$ from $0$ (no communication) to $50$ and record the result in Fig.~\ref{fig:hyper_M}. It is worth mentioning that we also test the performance of D-BCD after replacing the similarity-based model aggregation with mean aggregation, denoted by SC* and P-Pulse*. 
The classification performance of MNIST increases consistently as $M$ increases, which validates that D-SGD transfers knowledge across devices when local datasets on different devices obey the identical distribution. However, in a network with heterogenous on-device data distribution (SC and P-Pulse), the information from neighbors may potentially undermine ($3.72$\% average decline) the performance of local model, which can be observed when $M$ equals to $5$ and $10$. But when a larger number of neighbors are involved (i.e., $M=50$), the accuracy will rebound markedly, especially for D-BCD with similarity-based collaborative learning. 
Essentially, the higher $M$ is, the more likely the device can acquire knowledge from similar devices, therefore improving the local models. On the contrary, D-BCD without the similarity-based aggregation (dashed lines) achieves only limited performance improvement when $M$ increases.

Fig.~\ref{fig:hyper_mu} displays the test accuracy when $M=50$ and $\mu$ varies in $\{0.01, 0.1, 0.5, 0.9\}$. The variation of $\mu$ does not dramatically influence the performance on MNIST while increasing $\mu$ leads to inferior performance on SC and P-Pulse. This may also be caused by the noise coupled with the parameters passed from neighbors. For homogenous data distribution, all information from neighbors can be regarded as beneficial. But for the heterogenous one, the information from devices with different local data distributions may harm the model parameters learned from local data samples. In addition, a larger $\mu$ brings more noise to the final model, which results in the noticeable performance drop. According to Fig.~\ref{fig:hyper_communicaion}, D-BCD acts differently on homogenous and heterogenous data distributions. Meanwhile, inter-device communication is not always helpful for heterogenous data distribution. In this regard, identifying similar neighbors can improve the effectiveness of this collaborative learning process.

\begin{figure}[]
\centering
\subfigure[]{
\begin{minipage}[t]{0.5\linewidth}
\includegraphics[scale=0.51]{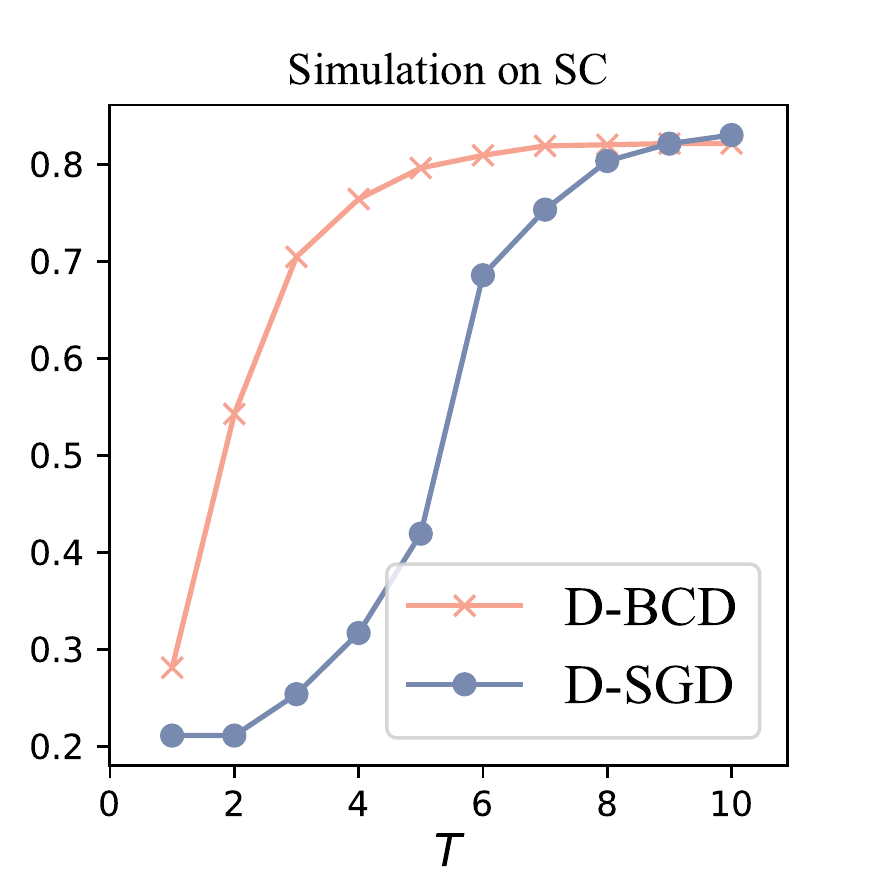}
\end{minipage}%
\label{fig:simulation1}
}%
\subfigure[]{
\begin{minipage}[t]{0.5\linewidth}
\includegraphics[scale=0.51]{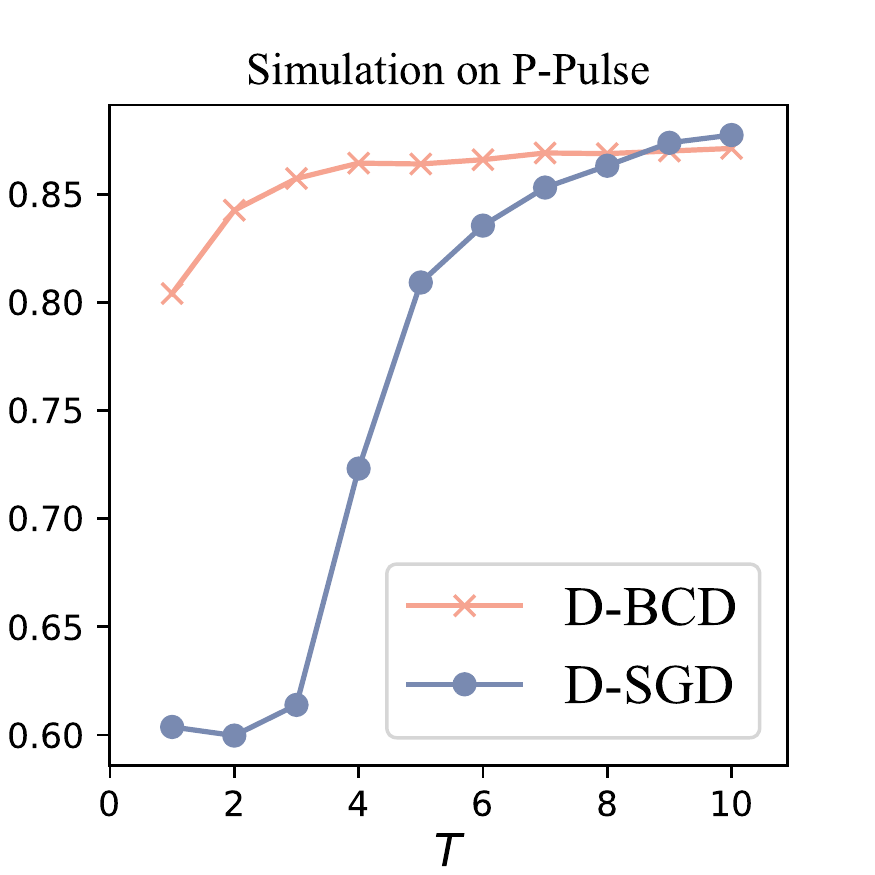}
\end{minipage}%
\label{fig:simulation2}
}%
\centering
\vspace{-0.15cm}
\caption{(a) and (b) reflect the fluctuation of the performance of D-BCD and D-SGD in the cold-start scenario and the limited communication scenario, respectively.}
\label{fig:simulation}
\vspace{0cm}
\end{figure}

\subsection{Simulation Study} \label{sec:Simulation}

Most smart devices are battery-powered, which means the allowed amount of inter-device communication is limited. 
Accordingly, we simulate a communication scenario where every device only communicates $60$ times per hour, i.e., once per $60s$.
Such communication frequency can improve the energy efficiency yet meets the requirement of most e-health monitoring scenarios.
Besides, we emulate the scenario where new local training data points emerge throughout the training process. This is done by gradually increasing the percentage of available training data $r\%$. Here we specify $r = T\times10\%$, where $T$ is the index of hour.
We assume that each personal dataset is from a $10$-hour overnight e-health monitoring, where the data size increments by $10$\% every hour. This is in line with the typical length of overnight recordings in SC and P-Pulse.
This setting is applied to all devices. We plot the overall test accuracy of the network of 10 hours' training using D-SGD and D-BCD in Fig. \ref{fig:simulation2}. Not surprisingly, D-BCD prevails over D-SGD while $T<9$ on both SC and P-Pulse, and the maximum performance differences are $45.06$\% and $24.30$\% on SC and P-Pulse, respectively. Thus, D-BCD is a promising learning framework that provides credible results for personalized E-health analytics, where the number of training iterations is strictly limited by the battery life of user devices. 

\section{Conclusion}
In this paper, we design and develop a novel learning framework, namely D-BCD for training personalized E-health analytic models in a fully decentralized and on-device manner. With the similarity-based collaborative learning scheme, D-BCD is able to achieve competitive performance compared with centralized peer methods, while being more efficient in training and resistant to gradient vanishing. Designed as a generic decentralized E-health framework, D-BCD can be utilized to optimize a wide range of predictive models across different tasks. The applicable datasets of D-BCD include domains like images, recommendation, time series, etc. Besides, the feasibility of D-BCD on two medical classification tasks -- sleep stage scoring (SC) and apnea detection (P-Pulse) -- has revealed its potential for generalization to other diseases. Further analyses on hyperparameters and scenario simulations suggest that D-BCD can alleviate cold-start issues and outperform SGD-based optimization framework when communication is limited. Such characteristics are of great importance to the on-device deployment environment, exhibiting the immense potential of D-BCD in E-health and telemedicine applications.

\balance
\section*{Acknowledgment}
This work is supported by Australian Research Council Future Fellowship (Grant No. FT210100624), Discovery Project (Grant No. DP190101985) and Discovery Early Career Research Award (Grant No. DE200101465).

\ifCLASSOPTIONcaptionsoff
	\newpage
\fi


\bibliographystyle{IEEEtran} 
\bibliography{jbhi} 

\end{document}